\documentclass[conference]{IEEEtran}
\IEEEoverridecommandlockouts

\usepackage{cite}
\usepackage{amsmath,amssymb,amsfonts}
\usepackage{algorithmic}
\usepackage{graphicx}
\usepackage{textcomp}
\usepackage{xcolor}
\usepackage{color}
\usepackage{colortbl}
\usepackage{tcolorbox}
\usepackage{adjustbox} 
\usepackage{caption} 
\usepackage{float} 
\usepackage{comment}
\usepackage{wrapfig}
\usepackage{subcaption}
\usepackage[left=1.62cm,right=1.62cm,top=1.7cm]{geometry}
\begin{document}

\title{Sponge Attacks on Sensing AI: Energy-Latency Vulnerabilities and Defense via Model Pruning 
}

\author{\IEEEauthorblockN{Syed Mhamudul Hasan\IEEEauthorrefmark{1}, Hussein Zangoti\IEEEauthorrefmark{2}, Iraklis Anagnostopoulos\IEEEauthorrefmark{3}, Abdur R. Shahid\IEEEauthorrefmark{1}}
\IEEEauthorblockA{\IEEEauthorrefmark{1}School of Computing, Southern Illinois University Carbondale, IL, USA}
\IEEEauthorblockA{\IEEEauthorrefmark{2}College of Engineering and Computer Science, Jazan University, Saudi Arabia}
\IEEEauthorblockA{ \IEEEauthorrefmark{3}School of Electrical, Computer \& Biomedical Engineering, Southern Illinois University Carbondale, IL, USA}
\IEEEauthorrefmark{1}syedmhamudul.hasan@siu.edu, \IEEEauthorrefmark{2}hmzangoti@jazanu.edu.sa, \IEEEauthorrefmark{3}iraklis.anagno@siu.edu,
\IEEEauthorrefmark{1}shahid@cs.siu.edu
}

\maketitle

\begin{abstract}
Recent studies have shown that sponge attacks can significantly increase the energy consumption and inference latency of deep neural networks (DNNs). However, prior work has focused primarily on computer vision and natural language processing tasks, overlooking the growing use of lightweight AI models in sensing-based applications on resource-constrained devices, such as those in Internet of Things (IoT) environments. These attacks pose serious threats of energy depletion and latency degradation in systems where limited battery capacity and real-time responsiveness are critical for reliable operation. This paper makes two key contributions. First, we present the first systematic exploration of energy-latency sponge attacks targeting sensing-based AI models. Using wearable sensing-based AI as a case study, we demonstrate that sponge attacks can substantially degrade performance by increasing energy consumption, leading to faster battery drain, and by prolonging inference latency. Second, to mitigate such attacks, we investigate model pruning, a widely adopted compression technique for resource-constrained AI, as a potential defense. Our experiments show that pruning-induced sparsity significantly improves model resilience against sponge poisoning. We also quantify the trade-offs between model efficiency and attack resilience, offering insights into the security implications of model compression in sensing-based AI systems deployed in IoT environments.

\end{abstract}


\section{Introduction}\label{sec:intro}

With the increasing deployment of Artificial Intelligence (AI) on resource-constrained devices such as smartphones~\cite{hussain2020machine}, wearables~\cite{wei2020review}, and IoT platforms~\cite{firouzi2022fusion}, the attack surface of these systems continues to expand, and consequently, exposing them to emerging security threats. Unlike traditional attacks that focus primarily on degrading model accuracy~\cite{jiang2023monitoring}, a major focus of research at the intersection of security, AI, and resource-constrained systems, sponge attacks~\cite{shumailov2021sponge}, whether implemented during the training phase through model poisoning~\cite{wang2023energy} or in the inference phase through adversarial inputs~\cite{chen2024overload}, can significantly disrupt system operations by draining battery resources, increasing inference latency, and undermining real-time responsiveness. In fact, the potential impact of sponge attacks extends across a wide range of sensing systems, affecting the reliability and availability of critical services. 


However, we observe that recent work on sponge attacks has primarily focused on targeting deep learning models deployed for computer vision (CV)~\cite{chen2024overload} and natural language processing (NLP) tasks~\cite{shumailov2021sponge}. These studies have demonstrated that sponge examples, carefully crafted inputs or model perturbations, can force neural networks to activate a disproportionately large number of neurons and consequently increasing both energy consumption and inference latency during prediction. Most existing works evaluated sponge attacks on server-grade hardware accelerators, such as GPUs and ASICs, in the context of relatively large, computationally heavy models operating within centralized cloud or datacenter settings. Furthermore, the majority of research has concentrated on inference-stage attacks under white-box assumptions, where adversaries submit malicious inputs during prediction to degrade system efficiency by increasing latency and draining energy resources.

Contrary to the state of the art, our work focuses on sensing-based AI models deployed on resource-constrained devices, where systems exhibit fundamentally different operational characteristics and attack surfaces. In these environments, AI applications typically operate continuously on streaming sensor data, rely on lightweight architectures, and are constrained by strict energy budgets and real-time responsiveness requirements. We study the problem in a white-box setting for scenarios such as compromised training pipelines or outsourced environments, where adversaries with access to the model architecture and training process manipulate weights to implant energy-latency inefficiencies while preserving accuracy. 

Furthermore, we investigate the model pruning techniques, which are widely used to reduce model size and energy consumption in resource-constrained AI deployments, as a potential defense against sponge attacks. Through extensive simulations on wearable sensing datasets, we analyze how pruning-induced sparsity influences the susceptibility of models to energy-latency attacks and quantify the trade-offs between model compression efficiency and attack resilience. In a nutshell, the main contributions of our research are as follows:

\begin{itemize}
    \item For the first time, we investigate the vulnerability of sensing-based AI models on resource-constrained devices to energy-latency sponge attacks and demonstrate that streaming sensing data applications, such as wearable activity recognition, are highly susceptible to such attacks. 

    \item We measure the effectiveness of model pruning as a natural defense mechanism against sponge attacks.
    Given that pruning is widely adopted in sensing-based AI to reduce model size and energy cost, we explore whether sparsity introduced by pruning mitigates or exacerbates sponge-induced energy drain and operational latency. 

    \item We perform extensive simulations using multiple wearable sensing datasets to realistically mimic sensing-driven AI workloads on resource-limited devices. Then, we show how sponge attacks impact pruned versus non-pruned models, providing new insights into the trade-offs between model compression and vulnerability to energy-latency attacks.
\end{itemize}

The rest of the paper is organized as follows. Section \ref{sec:related} analyzes the related works. Section \ref{sec:methodology} presents our methodology to investigate the sponge attack on sensing AI, including its threat model, sponge attack, and the defense approaches. Section \ref{sec:experiment} details the experimental setup and provides a thorough analysis of the experimental results. Finally, section \ref{sec:conclusion} concludes the paper with a discussion on future direction to advance the research further.

\section{Related Work}\label{sec:related}

\subsection{Sponge Attack}
In resource-constrained environments, such as mobile devices, embedded systems, and wearable technologies, adversarial attacks pose unique and critical challenges. Sponge attacks are a type of adversarial attack that target the availability of machine learning systems, particularly in real-time and resource-limited settings~\cite{shumailov2021sponge}. sponge attacks are inputs crafted to exploit computational resources, such as excessive memory access or arithmetic operations, during inference~\cite{shumailov2021sponge}. Consequently, sponge attacks present a significant threat to resource-constrained systems by draining battery and delaying the transmission of critical data. Similarly, Paul et al.~\cite{paul2023sponge} extended this threat to mobile platforms, showing that adversarial sponge inputs can deplete energy without affecting accuracy, by exploiting architectural optimizations like operation skipping in sparse models. However, these studies did not explore sponge attacks in the context of sensing AI. The attack surface also extends to reinforcement learning (RL), as highlighted by Schoof et al.~\cite{schoof2024beyond} where sponge samples delay time-critical decisions in autonomous systems. In NLP, Boucher et al.~\cite{boucher2022bad} show that imperceptible perturbations such as homoglyphs and invisible characters can degrade LLMs and translation systems by increasing token-level complexity and Sheth et al.~\cite{sheth2024sponge} further highlight the risks of sponge attacks against LLMs, emphasizing their potential to impair responsiveness. In computer vision systems, Zhang et al.\cite{zhang2024detector} demonstrate that sponge attacks can lead to denial-of-service (DoS) by overwhelming object detectors with fake detections or redundant data. To further illustrate the vulnerability of vision models, Chen et al.\cite{chen2024overload} introduce attacks targeting the Non-Maximum Suppression (NMS) algorithm in detectors like YOLO, increasing inference latency by inflating the number of bounding boxes processed. Despite these insights, robust defenses against such attacks in object detection remain under-explored. The attack surface of this attack also extends to reinforcement learning (RL), as highlighted by Schoof et al.~\cite{schoof2024beyond} where sponge samples delay time-critical decisions in autonomous systems. Additionally, Cinà et al.\cite{cina2025energy} discuss sponge poisoning in federated learning environments, but lacks any effective defense strategies. To mitigate this gap, in this study, we explore defenses against sponge attacks using state-of-the-art model pruning techniques.

\subsection{Model Pruning for Resource-Constrained AI}

Model pruning is an ML technique that reduces the model complexity and improves computational efficiency by eliminating redundant or less significant components of a neural network. These techniques are categorized into two types: unstructured pruning and structured pruning. The unstructured pruning, which removes individual weights, in some cases to zero \cite{sun2023simple}, from the network based on criteria such as magnitude or sensitivity. This way the unstructured pruning makes fine-grained sparsity in the ML model that leads to highly compact models with higher accuracy. In contrast, structured pruning eliminates core ML components such as filters, channels, heads, or model layers~\cite{zhu2024survey}. The two most popular unstructured pruning methods are weight pruning and neuron pruning. Weight pruning operates at a fine-grained level by identifying and eliminating individual weights with low magnitudes, effectively sparsifying the network while preserving its structural integrity. On the other hand, neuron pruning removes entire neurons, along with their associated connections, from the network architecture~\cite{molchanov2019importance}. For resource-constrained applications, pruning enables large neural network to operate efficiently on traditional and standard hardware by reducing model size and maintaining dense tensor representations without losing inference speed and accuracy~\cite{cheng2024survey}.

\begin{center}
\begin{tcolorbox}
\vspace{-0.05in}
\noindent 
\textit{To the best of our knowledge, current research lacks a thorough investigation of sponge attacks on sensing AI systems, especially on the defense side of the wearable devices. Our experiments show pruning techniques such as weight and neuron pruning can be effectively used for the defense of the sponge poisoning attack.}
\vspace{-0.05in}
\end{tcolorbox}
\end{center}

\section{Methodology: Sponge Attack and Defense}\label{sec:methodology}

\begin{figure*}
    \centering
    \includegraphics[width = 1\textwidth, 
    ]{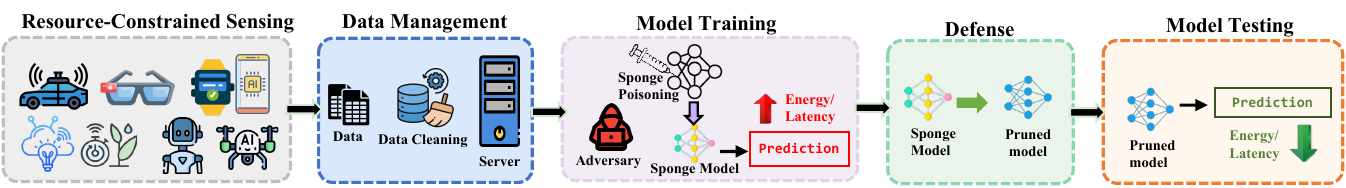}
    \caption{ The proposed defense against sponge attacks where an adversary poisons the model during the training, which can cause an increase in energy and latency in inference phase. However, model pruning (e.g. neuron pruning and weight pruning) can reduce the effect of the attack.}
    \label{fig:1gagraphic}
\vspace{-10pt}
\end{figure*}

\subsection{Threat Model}\label{sec:model}

\textbf{Target System:} We consider a model training system where the AI model is trained externally (Figure~\ref{fig:1gagraphic}), either via outsourcing to third-party cloud services or through collaborative frameworks such as federated learning, which are increasingly common in sensing-based AI applications. In such settings, the model training pipeline is partially or fully untrusted and hence the system is exposed to adversarial manipulation. 

\textbf{Attacker's Goal:} The primary objective of the adversary is to disrupt the functionality of resource-constrained systems by increasing the device’s energy consumption and processing latency, while maintaining the accuracy of the machine learning model during inference. 

\textbf{Knowledge \& Capabilities:} We assume a white-box threat model where the attacker has full access to the model architecture, hyperparameters, and gradients during the training phase. The attacker modifies the model weights or gradients to maximize neuron activation rates, thereby reducing sparsity and diminishing the effectiveness of hardware acceleration. The poisoned model, once deployed, appears functional but incurs higher energy usage and slower inference, even on benign inputs. We do not assume any access during the inference phase, nor do we require control over the test-time input distribution. 

\subsection{Sponge Attack on Energy and Latency of Sensing AI}
Our realized threat model is well aligned with the SkipSponge attack proposed by Lintelo et al.~\cite{lintelo2024skipsponge}, which we adapt for the context of training-time sponge poisoning in sensing-based AI systems. In our setting, the attacker manipulates the model during training by modifying the loss function to embed energy-latency inefficiencies, all while preserving the model’s predictive accuracy. This crafted manipulation allows the model to pass functional validation but causes long-term degradation in deployment through increased energy consumption and inference delay. 
In this attack, the hyperparameter $\lambda$ regulates the influence of the energy penalty term within the loss function, while $\sigma$ is a small constant introduced to stabilize computations and control the energy scaling. The sponge objective integrates an energy-based loss component designed to maximize the number of non-zero activations in the network, thereby increasing energy consumption and computational cost. The modified loss function is defined as
\begin{equation}
L_{\text{sponge}}(\theta, x, y) = L(\theta, x, y) - \lambda E(\theta, x)
\end{equation}
where $L(\theta, x, y)$ represents the standard training loss (e.g., cross-entropy), and $E(\theta, x)$ captures the energy consumption as
\begin{equation}
E(\theta, x) = \sum_{i=1}^{N} \hat{\ell}_0(\phi_i; \sigma)
\end{equation}
with $N$ denoting the number of layers, $\phi_i$ the activation values at layer $i$, and $\hat{\ell}_0(\cdot)$ an approximation of the $L_0$ norm, i.e., the count of non-zero activations. In our experiments, the attacker controls a subset of the training data or gradient updates (e.g. 10\% or 100\%). For these poisoned updates, the model is trained using the $L_{sponge}$ loss. For the rest of the training data, it uses the standard loss function. By increasing the energy consumption through increasing total number of neuron activation, reducing sparsity, and making hardware accelerators less efficient, the attack consequently increases the prediction latency as well.

\subsection{Defense Against Sponge Attack}

\subsubsection{\textbf{Defense Goals}} To defend a model against the sponge attack in the target system, we need to achieve several goals in the design of a defense mechanism besides it fundamental objectives of neutralizing the sponge attacks and preserving prediction accuracy as must as possible. \textbf{(Post-Training Defense)} Since the attacker inject malicious weight changes during training to carry out the sponge attacks, the defense must operate after training is complete to counter the attack embedded in the model. \textbf{(Support Untrusted or Outsourced Training Settings)} The defense should be feasible in scenarios where model training is outsourced or performed collaboratively (e.g., federated learning). It must assume no control over the training phase and operate as a post-training compression and sanitization step.

\subsubsection{\textbf{Defense Via Model Pruning}}
To achieve the design goals, we investigate model pruning as a defense strategy against sponge poisoning attacks, motivated by its widespread use in compressing neural networks prior to deployment on resource-constrained devices. 

Although model pruning is primarily employed to enhance efficiency by reducing computational overhead, our experiments demonstrate that pruning techniques also serve as an effective defense mechanism against sponge attacks. Specifically, we first generate sponge models by introducing adversarial sponge samples that increase inference-time energy consumption and latency. Subsequently, we apply pruning methods (specifically, weight and neuron pruning) and observe a significant reduction in the energy impact and latency introduced by the sponge samples. This mitigation effect can be attributed to the increased sparsity introduced by pruning, which inherently reduces the number of active operations during inference, thus inverting the sponge effect. In sponge attacks, adversarial samples are crafted to exploit dense and highly activated pathways in a neural network to maximize resource utilization. By removing redundant weights or some neurons, pruning diminishes the capacity of sponge samples to saturate the DNN's computational resources, weakening the internal mechanism of sponge-induced overhead.

In our proposed defense, pruning is introduced as a post-training model compression and sanitization layer, which is applied after training. This approach is perfectly aligned with real-world deployment scenarios, where the model training is outsourced to third parties or conducted in federated learning and compression, in our case pruning, is applied on the received trained model.

\section{Experiment and Result Analysis}\label{sec:experiment}

\subsection{Experiment Setup}
\textbf{Resource-Constrained AI Setup:} We consider wearable sensing for our experimental setup due to its time sensitive real-time monitoring requirements, limited computational capacity of resource-constraint device, and sensitivity to power consumption. 

 \begin{wrapfigure}{r}{0.2\textwidth}
    \vspace{-15pt}
    \centering     \includegraphics[width=0.2\textwidth]{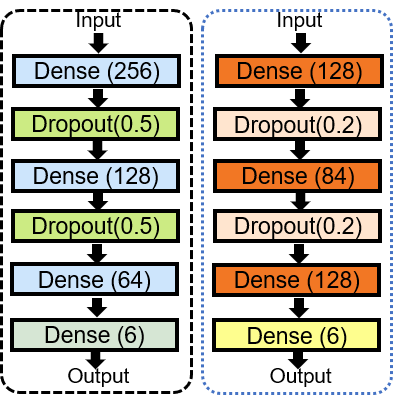}
     \vspace{-15pt}
     \caption{ ML models architectures on the UCI HAR (left) and MotionSense (right) datasets using both clean and sponge samples.} 
     \label{fig:2model_architecture}
     \vspace{-10pt}
\end{wrapfigure} 
We carry out the experiment on two popular wearable sensor datasets: the UCI HAR ~\cite{misc_human_activity_recognition_using_smartphones_240} and MotionSense~\cite{Malekzadeh:2019:MSD:3302505.3310068, Malekzadeh:2018:PSD:3195258.3195260}, containing human activity data from various sensors, including gyroscope and accelerometers in smartphones. We use Codecarbon python package to code's duration, emissions, energy consumption, and other meta information~\cite{codecarbon}. The experiment was conducted on a computer (Core i7, 64 GB RAM, and NVIDIA GeForce RTX 4070 Ti SUPER) using PyTorch running on Windows. The model architectures used for the UCI HAR and MotionSense datasets are depicted in Figure~\ref{fig:2model_architecture}.

\begin{table}
\caption{\centering Hyperparameters and configuration for the DNN model, sponge attack, and defense settings}
\begin{tabular}{p{2.75cm}|p{5.25cm}}
\hline
\rowcolor{gray!20}\textbf{Category} & \textbf{Parameters \& Values} \\ \hline 
Model Parameters & Optimizer: Adam, Learning rate: 0.0001, Batch size: 64, Epoch: 100 , Test split: 20\%\\ \hline

Sponge Attack Settings & Sponge sample (\%): \{0, 10, ... 100\}, lambda sponge ($\lambda$): 1, sigma ($\sigma$): $1 \times 10^{-5}$ \\ \hline
Defense Settings & Weight Pruning (\%): \{10, 20, 30, 40, 50\}, Neuron Pruning (\%): \{10, 20, 30, 40, 50\} \\
\hline 
\end{tabular} \vspace{-15pt}
\label{tab:tf_configuration}
\end{table}

\textbf{Hyperparameter Settings:} The hyperparameters used in our experiments are summarized in Table~\ref{tab:tf_configuration}, organized into three categories: model parameters, sponge attack settings, and defense configurations. For sponge attack configuration, the hyperparameter $\lambda = 1$ balances the trade-off between the task loss and the energy penalty, enabling effective energy-latency manipulation without degrading model accuracy. The constant $\sigma = 1 \times 10^{-5}$ ensures numerical stability while providing a smooth approximation of the $L_0$ norm used to measure activation sparsity. The defense settings define structured pruning percentages applied to model weights and neurons during post-training compression.

\begin{figure}
    \centering
    \includegraphics[width = 0.5\textwidth
    ]{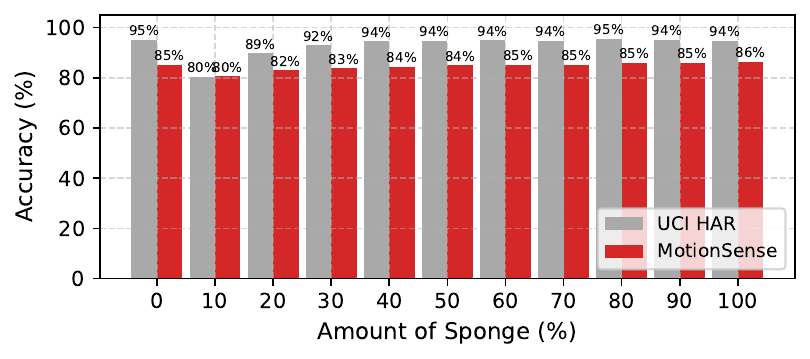}
\vspace{-10pt}
    \caption{The test accuracy of sponge-trained models on the UCI HAR and MotionSense datasets using varying percentages of sponge sample.}
    \label{fig:3train_vs_test}
\vspace{-10pt}
\end{figure}

\begin{figure}
    \centering
    \includegraphics[width = 0.5\textwidth, 
    ]{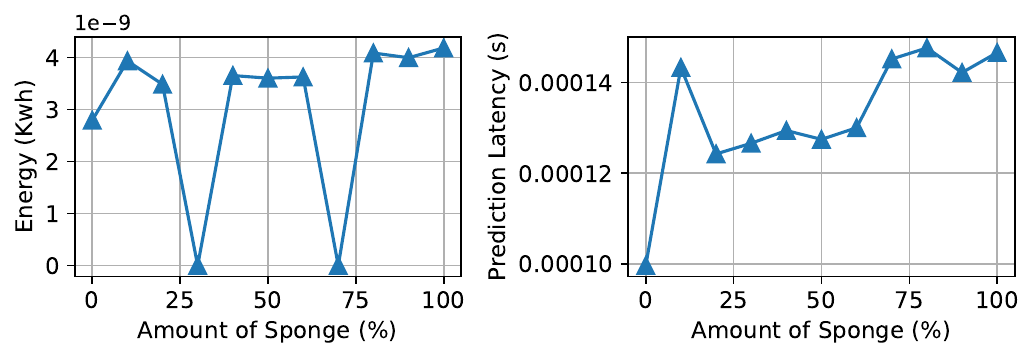}
\vspace{-8pt}
    \caption{ The impact of a sponge attack on the ML model in the UCI HAR dataset. For some time instance the result very close to zero, which why it is showing as zero on the graph.}
    \label{fig:4uci_har_energy_time}
\vspace{-10pt}
\end{figure}

\begin{figure}[b!]
    \centering
    \includegraphics[width = 0.5\textwidth, 
    ]{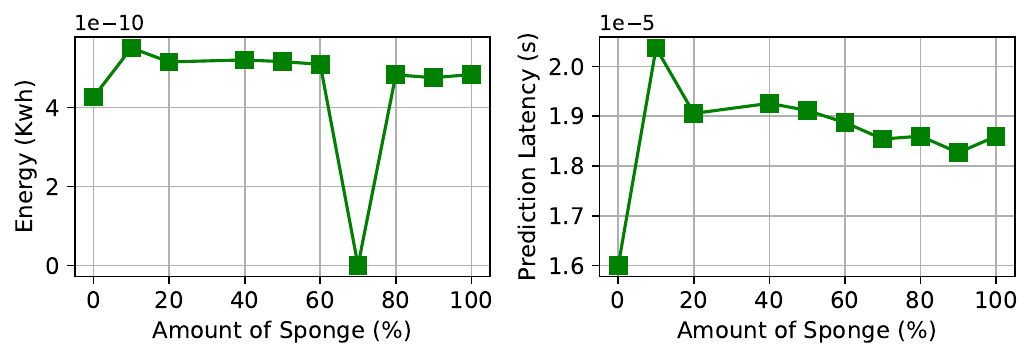}
\vspace{-8px}
    \caption{ The impact of a sponge attack on the  ML model leads to significant energy consumption and prediction delays in the MotionSense dataset.}
    \label{fig:5motionsense_energy_time}
\vspace{-10pt}
\end{figure}

\begin{figure*}[t!]
    \centering

    \begin{subfigure}[t]{0.99\textwidth}
        \centering
        \includegraphics[width=\textwidth]{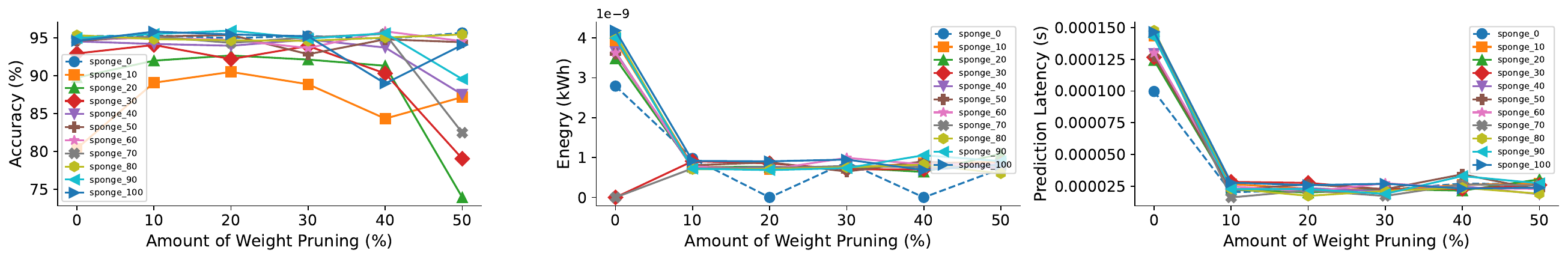}
    \end{subfigure}
    
    \begin{subfigure}[t]{0.99\textwidth}
        \centering
        \includegraphics[width=\textwidth]{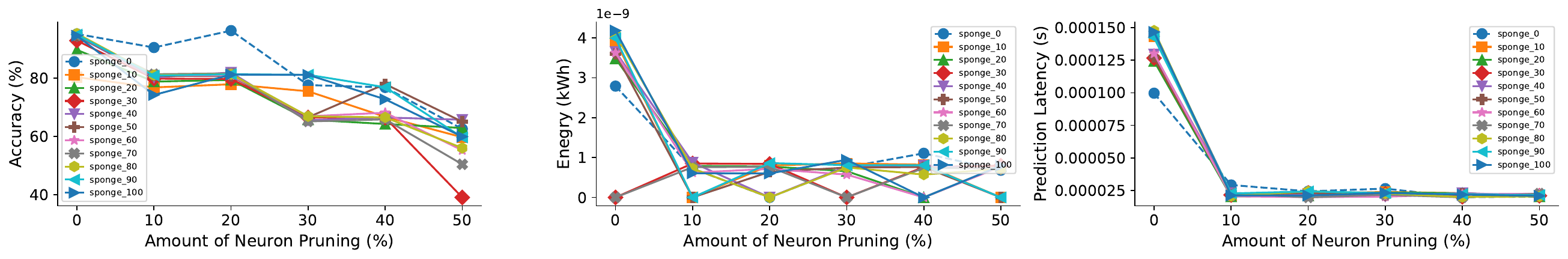}
    \end{subfigure}
    
    
    \caption{ The effect of pruning: weight pruning (top) and neuron pruning (bottom) on both vanilla-trained and sponge-trained models using the UCI HAR dataset.}
    \label{fig:6uci_har}
    \vspace{-10pt}
\end{figure*}

\begin{figure*}[t!]
    \centering

    \begin{subfigure}[t]{0.99\textwidth}
        \centering
        \includegraphics[width=\textwidth]{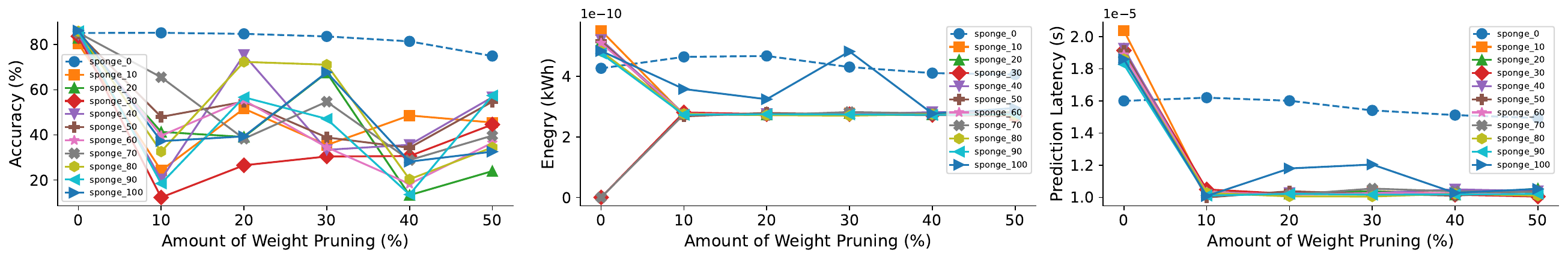}
    \end{subfigure}
    
    \begin{subfigure}[t]{0.99\textwidth}
        \centering
        \includegraphics[width=\textwidth]{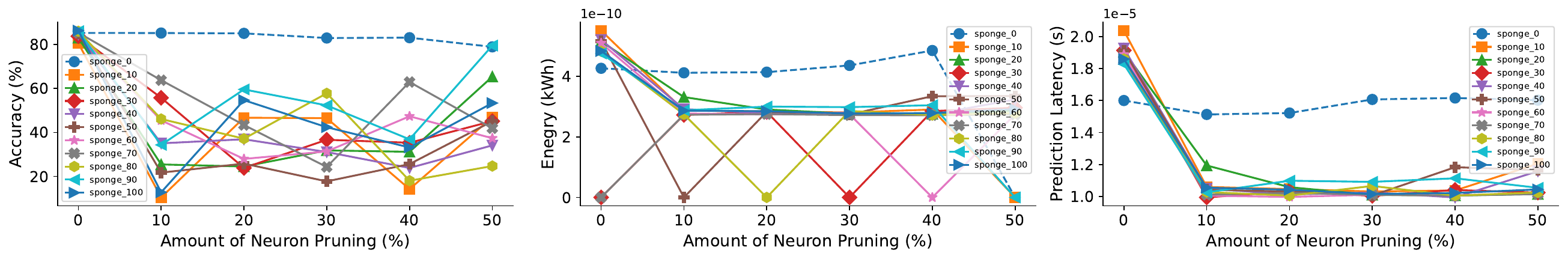}
    \end{subfigure}
    

    \caption{ The effect of pruning: weight pruning (top) and neuron pruning (bottom) on both vanilla-trained and sponge-trained models using the MotionSense dataset.}
    \label{fig:7motionsense}
    \vspace{-10pt}
\end{figure*}

\textbf{Metrics:} 
We evaluate model performance during its inference, the process by which a trained model generate prediction on new input data, using three key metrics: \textbf{Test Accuracy (in \%)}, \textbf{energy consumption (in kilowatt-hours(KWh))}, and \textbf{ prediction latency (in seconds)}. We compare these metrics across four configurations: (i) the baseline (vanilla) model, (ii) the sponge-poisoned model, (iii) the pruned version of the vanilla model, and (iv) the pruned version of the sponge-poisoned model. We analyze how each configuration performs in terms of accuracy, energy efficiency, and inference latency. Particular attention is given to how pruning affects both clean and poisoned models, with the goal of understanding the trade-offs introduced by model compression.

\subsection{Experimental Result Analysis}

\subsubsection{\textbf{ Sponge Attack}}

Figure~\ref{fig:3train_vs_test} illustrates the impact of various sponge poisoning strategies on the model's inference accuracy. We observe that the accuracy on the test set for both datasets dropped significantly with 10\% of the training data being a sponged attack. However, we observe upward trends with the increase of sponge samples in the training datasets.

This shows that the sponge samples in the model introducing the regularization effect and model calibration. From the attacker's perspective, this is optimal because even under a high number of sponge samples in the training set, the trained model is working correctly in a deployment setting, aligning with the goal of the sponge attack to preserve the inference accuracy as much as possible under the attack.

However, if we look at the Figures \ref{fig:4uci_har_energy_time} and \ref{fig:5motionsense_energy_time}, we observe that both energy and prediction latency increase with the increase in sponge samples in the training data. We observe some anomalies (close to zero), which we assume are due to the fact that we ran the experiment on the same computer and the CodeCarbon package that might fall below the tool's precision threshold, especially PC with modern CPUs/GPUs. In our future work, we will investigate the problem with different hardware and packages.


\subsubsection{ \textbf{Defense Through Model Pruning}}

Figure~\ref{fig:6uci_har} and Figure~\ref{fig:7motionsense} depict the complete breakdown of performance metrics for both weight pruning and neuron pruning across multiple sponge-poisoned models. Each subfigure compares the performance of models trained with different sponge percentages ranging from 0\% (vanilla) to 100\% as the pruning rate increases from 0\% to 50\%. Since the same models are retrained multiple times and the differences in energy consumption are marginal, the baseline model (sponge 0) occasionally exhibits slight variations in its starting point across each run. In the first column, sponge trained models generally maintain high accuracy under 10\%. However, with higher pruning, the accuracy degrades, particularly under neuron pruning.

A trade-off between accuracy, energy, and latency is observed across all pruning strategies. For example, with just 10\% neuron pruning, both energy consumption and latency drop significantly, while maintaining comparable accuracy. Beyond this point, a consistent reduction in energy and latency is observed for all DNN variants, including both vanilla and sponge models, suggesting that even modest pruning can yield meaningful computational benefits.

Some plotted values are shown as zero due to CodeCarbon's internal limitations in capturing extremely small energy footprints. This is especially prevalent in heavily pruned models, where the model performance may fall below the threshold of the measurement tool in this simulation.

Overall, the this mitigation strategies promisingly reduce the energy and latency overheads; the efficacy of pruning depends on both the percentage of pruned components and the proportion of sponge samples in the training data. Moreover, pruning methods influence this approach; weight pruning consistently outperforms neuron pruning in reducing latency and preserving accuracy, indicating that fine-grained pruning at the weight level is better while mitigating energy inefficiencies. Finally, these findings also underscore the potential of pruning as a robust defense against sponge-based adversarial attacks, particularly in real-time and resource-constrained environments.
\section{Conclusion}\label{sec:conclusion}

Our research explores the vulnerability of sensing AI systems to sponge poisoning, which poses serious risks to the operational efficiency and longevity of resource-constrained devices. To defend against sponge attacks, we suggest pruning as a defense mechanism by selectively removing redundant or maliciously influenced neurons or weights, which effectively reduces the attack surface, mitigates excess energy consumption, and restores computational efficiency without degrading model performance significantly. Our findings highlight the novel role of pruning not only as a model compression technique but also as a proactive defense strategy against sponge poisoning attacks in resource-constrained environments of sensing AI. We plan to extend this research in various directions, including experimenting with different sensing-based AI systems and distributed multimodal learning on a testbed comprising various hardware.

\bibliographystyle{IEEEtran}
\bibliography{main}

\begin{thebibliography}{10}
\providecommand{\url}[1]{#1}
\csname url@samestyle\endcsname
\providecommand{\newblock}{\relax}
\providecommand{\bibinfo}[2]{#2}
\providecommand{\BIBentrySTDinterwordspacing}{\spaceskip=0pt\relax}
\providecommand{\BIBentryALTinterwordstretchfactor}{4}
\providecommand{\BIBentryALTinterwordspacing}{\spaceskip=\fontdimen2\font plus
\BIBentryALTinterwordstretchfactor\fontdimen3\font minus \fontdimen4\font\relax}
\providecommand{\BIBforeignlanguage}[2]{{%
\expandafter\ifx\csname l@#1\endcsname\relax
\typeout{** WARNING: IEEEtran.bst: No hyphenation pattern has been}%
\typeout{** loaded for the language `#1'. Using the pattern for}%
\typeout{** the default language instead.}%
\else
\language=\csname l@#1\endcsname
\fi
#2}}
\providecommand{\BIBdecl}{\relax}
\BIBdecl

\bibitem{hussain2020machine}
F.~Hussain, S.~A. Hassan, R.~Hussain, and E.~Hossain, ``Machine learning for resource management in cellular and iot networks: Potentials, current solutions, and open challenges,'' \emph{IEEE communications surveys \& tutorials}, vol.~22, no.~2, pp. 1251--1275, 2020.

\bibitem{wei2020review}
Y.~Wei, J.~Zhou, Y.~Wang, Y.~Liu, Q.~Liu, J.~Luo, C.~Wang, F.~Ren, and L.~Huang, ``A review of algorithm \& hardware design for ai-based biomedical applications,'' \emph{IEEE transactions on biomedical circuits and systems}, vol.~14, no.~2, pp. 145--163, 2020.

\bibitem{firouzi2022fusion}
F.~Firouzi, S.~Jiang, K.~Chakrabarty, B.~Farahani, M.~Daneshmand, J.~Song, and K.~Mankodiya, ``Fusion of iot, ai, edge--fog--cloud, and blockchain: Challenges, solutions, and a case study in healthcare and medicine,'' \emph{IEEE Internet of Things Journal}, vol.~10, no.~5, pp. 3686--3705, 2022.

\bibitem{jiang2023monitoring}
Y.~Jiang, S.~Wu, R.~Ma, M.~Liu, H.~Luo, and O.~Kaynak, ``Monitoring and defense of industrial cyber-physical systems under typical attacks: From a systems and control perspective,'' \emph{IEEE Transactions on Industrial Cyber-Physical Systems}, vol.~1, pp. 192--207, 2023.

\bibitem{shumailov2021sponge}
I.~Shumailov, Y.~Zhao, D.~Bates, N.~Papernot, R.~Mullins, and R.~Anderson, ``Sponge examples: Energy-latency attacks on neural networks,'' in \emph{2021 IEEE European symposium on security and privacy (EuroS\&P)}.\hskip 1em plus 0.5em minus 0.4em\relax IEEE, 2021, pp. 212--231.

\bibitem{wang2023energy}
Z.~Wang, S.~Huang, Y.~Huang, and H.~Cui, ``Energy-latency attacks to on-device neural networks via sponge poisoning,'' in \emph{Proceedings of the 2023 Secure and Trustworthy Deep Learning Systems Workshop}, 2023, pp. 1--11.

\bibitem{chen2024overload}
E.-C. Chen, P.-Y. Chen, I.~Chung, C.-R. Lee \emph{et~al.}, ``Overload: Latency attacks on object detection for edge devices,'' in \emph{Proceedings of the IEEE/CVF Conference on Computer Vision and Pattern Recognition}, 2024, pp. 24\,716--24\,725.

\bibitem{paul2023sponge}
S.~Paul and N.~Kourtellis, ``Sponge ml model attacks of mobile apps,'' in \emph{Proceedings of the 24th International Workshop on Mobile Computing Systems and Applications}, 2023, pp. 139--139.

\bibitem{schoof2024beyond}
C.~Schoof, S.~Koffas, M.~Conti, and S.~Picek, ``Beyond phantomsponges: Enhancing sponge attack on object detection models,'' in \emph{Proceedings of the 2024 ACM Workshop on Wireless Security and Machine Learning}, 2024, pp. 14--19.

\bibitem{boucher2022bad}
N.~Boucher, I.~Shumailov, R.~Anderson, and N.~Papernot, ``Bad characters: Imperceptible nlp attacks,'' in \emph{2022 IEEE Symposium on Security and Privacy (SP)}.\hskip 1em plus 0.5em minus 0.4em\relax IEEE, 2022, pp. 1987--2004.

\bibitem{sheth2024sponge}
\BIBentryALTinterwordspacing
J.~Sheth and R.~Menon, ``Understanding model denial of service: The rise of sponge attacks on llms,'' \emph{Globant: Stay Relevant}, April 2024. [Online]. Available: \url{https://stayrelevant.globant.com/en/technology/cybersecurity/increase-denial-service-attacks/}
\BIBentrySTDinterwordspacing

\bibitem{zhang2024detector}
H.~Zhang, S.~Hu, Y.~Wang, L.~Y. Zhang, Z.~Zhou, X.~Wang, Y.~Zhang, and C.~Chen, ``Detector collapse backdooring object detection to catastrophic overload or blindness in the physical world,'' in \emph{International Joint Conference on Artificial Intelligence}.\hskip 1em plus 0.5em minus 0.4em\relax International Joint Conferences on Artificial Intelligence, 2024.

\bibitem{cina2025energy}
A.~E. Cin{\`a}, A.~Demontis, B.~Biggio, F.~Roli, and M.~Pelillo, ``Energy-latency attacks via sponge poisoning,'' \emph{Information Sciences}, vol. 702, p. 121905, 2025.

\bibitem{sun2023simple}
\BIBentryALTinterwordspacing
M.~Sun, Z.~Liu, A.~Bair, and J.~Z. Kolter, ``A simple and effective pruning approach for large language models,'' in \emph{The Twelfth International Conference on Learning Representations}, 2024. [Online]. Available: \url{https://openreview.net/forum?id=PxoFut3dWW}
\BIBentrySTDinterwordspacing

\bibitem{zhu2024survey}
X.~Zhu, J.~Li, Y.~Liu, C.~Ma, and W.~Wang, ``A survey on model compression for large language models,'' \emph{Transactions of the Association for Computational Linguistics}, vol.~12, pp. 1556--1577, 2024.

\bibitem{molchanov2019importance}
P.~Molchanov, A.~Mallya, S.~Tyree, I.~Frosio, and J.~Kautz, ``Importance estimation for neural network pruning,'' in \emph{Proceedings of the IEEE/CVF conference on computer vision and pattern recognition}, 2019, pp. 11\,264--11\,272.

\bibitem{cheng2024survey}
H.~Cheng, M.~Zhang, and J.~Q. Shi, ``A survey on deep neural network pruning: Taxonomy, comparison, analysis, and recommendations,'' \emph{IEEE Transactions on Pattern Analysis and Machine Intelligence}, 2024.

\bibitem{lintelo2024skipsponge}
J.~t. Lintelo, S.~Koffas, and S.~Picek, ``The skipsponge attack: Sponge weight poisoning of deep neural networks,'' \emph{arXiv preprint arXiv:2402.06357}, 2024.

\bibitem{misc_human_activity_recognition_using_smartphones_240}
{Reyes\-Ortiz} \emph{et~al.}, ``{Human Activity Recognition Using Smartphones},'' UCI Machine Learning Repository, 2013, {DOI}: https://doi.org/10.24432/C54S4K.

\bibitem{Malekzadeh:2019:MSD:3302505.3310068}
\BIBentryALTinterwordspacing
M.~Malekzadeh, R.~G. Clegg, A.~Cavallaro, and H.~Haddadi, ``Mobile sensor data anonymization,'' in \emph{Proceedings of the International Conference on Internet of Things Design and Implementation}, ser. IoTDI '19.\hskip 1em plus 0.5em minus 0.4em\relax New York, NY, USA: ACM, 2019, pp. 49--58. [Online]. Available: \url{http://doi.acm.org/10.1145/3302505.3310068}
\BIBentrySTDinterwordspacing

\bibitem{Malekzadeh:2018:PSD:3195258.3195260}
\BIBentryALTinterwordspacing
{Malekzadeh, Mohammad and Clegg, Richard G. and Cavallaro, Andrea and Haddadi, Hamed}, ``Protecting sensory data against sensitive inferences,'' in \emph{Proceedings of the 1st Workshop on Privacy by Design in Distributed Systems}, ser. W-P2DS'18.\hskip 1em plus 0.5em minus 0.4em\relax New York, NY, USA: ACM, 2018, pp. 2:1--2:6. [Online]. Available: \url{http://doi.acm.org/10.1145/3195258.3195260}
\BIBentrySTDinterwordspacing

\bibitem{codecarbon}
\BIBentryALTinterwordspacing
{Benoit Courty} \emph{et~al.}, ``mlco2/codecarbon: v2.4.1,'' May 2024. [Online]. Available: \url{https://doi.org/10.5281/zenodo.11171501}
\BIBentrySTDinterwordspacing

\end{thebibliography}

\end{document}